\documentclass[sigconf]{acmart}

\AtBeginDocument{%
  \providecommand\BibTeX{{%
    \normalfont B\kern-0.5em{\scshape i\kern-0.25em b}\kern-0.8em\TeX}}}

\copyrightyear{2024}
\acmYear{2024}
\setcopyright{rightsretained}
\acmConference[WWW '24] {Proceedings of the ACM Web Conference 2024}{May 13--17, 2024}{Singapore, Singapore.}
\acmBooktitle{Proceedings of the ACM Web Conference 2024 (WWW '24), May 13--17, 2024, Singapore, Singapore}
\acmPrice{}
\acmISBN{979-8-4007-0171-9/24/05}
\acmDOI{10.1145/XXXXXX.XXXXXX}

\begin{document}

\title{MentaLLaMA: Interpretable Mental Health Analysis on Social Media with Large Language Models}

\author{Kailai Yang}
\affiliation{%
  \institution{The University of Manchester}
  \city{Manchester}
  \country{United Kingdom}
}
\email{kailai.yang@manchester.ac.uk}

\author{Tianlin Zhang}
\affiliation{%
  \institution{The University of Manchester}
  \city{Manchester}
  \country{United Kingdom}
}
\email{tianlin.zhang@postgrad.manchester.ac.uk}

\author{Ziyan Kuang}
\affiliation{%
  \institution{Jiangxi Normal University}
  \city{Nanchang}
  \country{China}
}
\email{202340101007@jxnu.edu.cn}

\author{Qianqian Xie}
\authornote{Corresponding author. Qianqian is now affiliated with Yale University.}
\affiliation{%
  \institution{The University of Manchester}
  \city{Manchester}
  \country{United Kingdom}
}
\email{xqq.sincere@gmail.com}

\author{Jimin Huang}
\affiliation{%
  \institution{Wuhan University}
  \city{Wuhan}
  \country{China}
}
\email{jimin@chancefocus.com}

\author{Sophia Ananiadou}
\affiliation{%
  \institution{The University of Manchester}
  \city{Manchester}
  \country{United Kingdom}
}
\email{sophia.ananiadou@manchester.ac.uk}

\renewcommand{\shortauthors}{Kailai Yang et al.}

\begin{abstract}
As an integral part of people’s daily lives, social media is becoming a rich source for automatic mental health analysis. As traditional discriminative methods bear poor generalization ability and low interpretability, the recent large language models (LLMs) have been explored for interpretable mental health analysis on social media, which aims to provide detailed explanations along with predictions in zero-shot or few-shot settings. The results show that LLMs still achieve unsatisfactory classification performance in a zero-shot/few-shot manner, which further significantly affects the quality of the generated explanations. Domain-specific finetuning is an effective solution, but faces two critical challenges: 1) lack of high-quality training data. 2) no open-source foundation LLMs. To alleviate these problems, we formally model interpretable mental health analysis as a text generation task, and build the first multi-task and multi-source interpretable mental health instruction (IMHI) dataset with 105K data samples to support LLM instruction tuning and evaluation. The raw social media data are collected from 10 existing sources covering 8 mental health analysis tasks. We prompt ChatGPT with expert-designed few-shot prompts to obtain explanations. To ensure the reliability of the explanations, we perform strict automatic and human evaluations on the correctness, consistency, and quality of generated data. Based on the IMHI dataset and LLaMA2 foundation models, we train MentaLLaMA, the first open-source instruction-following LLM series for interpretable mental health analysis on social media. We evaluate MentaLLaMA and other advanced methods on the IMHI benchmark, the first holistic evaluation benchmark for interpretable mental health analysis. The results show that MentaLLaMA approaches state-of-the-art discriminative methods in correctness and generates human-level explanations. MentaLLaMA models also show strong generalizability to unseen tasks. The project is available at \url{https://github.com/SteveKGYang/MentaLLaMA}.
\end{abstract}


\keywords{mental health analysis, interpretability, social media, large language models}



\maketitle

\section{Introduction}
Mental health-related issues are posing increasing threats to public health worldwide~\cite{evans2018socio}, but remain underestimated due to the lack of social awareness and stigma~\cite{pirina2018identifying}. With the development of web technology, social media has become an integral part of people's daily lives\footnote{\url{https://wearesocial.com/uk/blog/2022/01/digital-2022/}}. Many people with potential mental health issues turn to social media platforms such as Twitter and Reddit to share their feelings, which makes social media texts a rich source for mental health analysis and potential early intervention~\cite{balani2015detecting,tsakalidis2019can}. However, manual mental health analysis on social media becomes impossible with the explosive amounts of social media posts. Therefore, many works explore natural language processing (NLP) techniques to perform automatic mental health analysis on social media~\cite{garg2023mental}.

In NLP for mental health, previous methods mainly model mental health analysis on social media as text classification tasks, where pre-trained language models (PLMs)~\cite{ji-etal-2022-mentalbert} achieve state-of-the-art (SOTA) performance. However, PLMs often struggle with poor generalization to unseen tasks and lack of robustness in multi-task scenarios~\cite{wang2022language,moor2023foundation}. Another key limitation of these methods is that they make discriminative predictions with low interpretability, limiting their reliability in practical usage. To alleviate these problems, the latest large language models (LLMs), such as ChatGPT\footnote{\url{https://openai.com/blog/chatgpt}} and GPT-4~\cite{OpenAI2023GPT4TR}, are explored~\cite{yang2023evaluations,xu2023leveraging} on detecting multiple mental health conditions and providing detailed explanations for their decisions, because they are proven to demonstrate superior generalization capabilities~\cite{brown2020language,wei2022emergent}. 
Specifically, Yang et al.~\cite{yang2023evaluations} performed comprehensive study and careful human evaluations to show that ChatGPT has strong in-context learning ability and can generate approaching-human explanations for its correct classifications, indicating its potential to enhance the interpretability of mental health analysis.

However, closed-source LLMs such as ChatGPT still struggle to achieve comparable mental health classification performance to SOTA supervised methods in a zero-shot~\cite{amin2023will} or few-shot~\cite{xu2023leveraging} learning setting. Moreover, such low precision is proven to further significantly affect the quality of the generated explanations, known as inaccurate reasoning~\cite{yang2023evaluations}. An effective solution is to fine-tune LLMs with task-specific data~\cite{han2023medalpaca,xie2023pixiu}, which can better align LLMs with the target domain while still keeping strong generalization ability. However, there are two key challenges in improving LLMs for interpretable mental health analysis with fine-tuning. Firstly, fine-tuning LLMs requires high-quality supervised training data. In mental health analysis on social media, though a few datasets include short extracted casual text spans~\cite{garg-etal-2022-cams,garg-etal-2023-annotated}, it still lacks open-source data that provides detailed and reliable explanations for detection results. This is mainly due to the sensitive research topic~\cite{nicholas2020ethics,benton2017ethical} and the high cost of writing explanations by domain experts. Secondly, prompting or fine-tuning close-source LLMs such as ChatGPT can be expensive\footnote{\url{https://openai.com/pricing}}, time-consuming, and with huge carbon emissions\footnote{\url{https://www.cutter.com/article/environmental-impact-large-language-models}}, while no open-source LLMs for interpretable mental health analysis have been released for public use. The lack of resources and high costs hinder the progress in related research.

To bridge these gaps, we formally model interpretable mental health analysis as a text-generation task, which aims to detect evidence of mental health conditions on social media posts and generate explanations for the predictions.
We build the first multi-task and multi-source \textbf{I}nterpretable \textbf{M}ental \textbf{H}ealth \textbf{I}nstruction (IMHI) dataset with 105K data samples to support LLM instruction tuning~\cite{ouyang2022training} and evaluation. 

\begin{figure}[htpb]
\centering
\includegraphics[width=8cm,height=5.953cm]{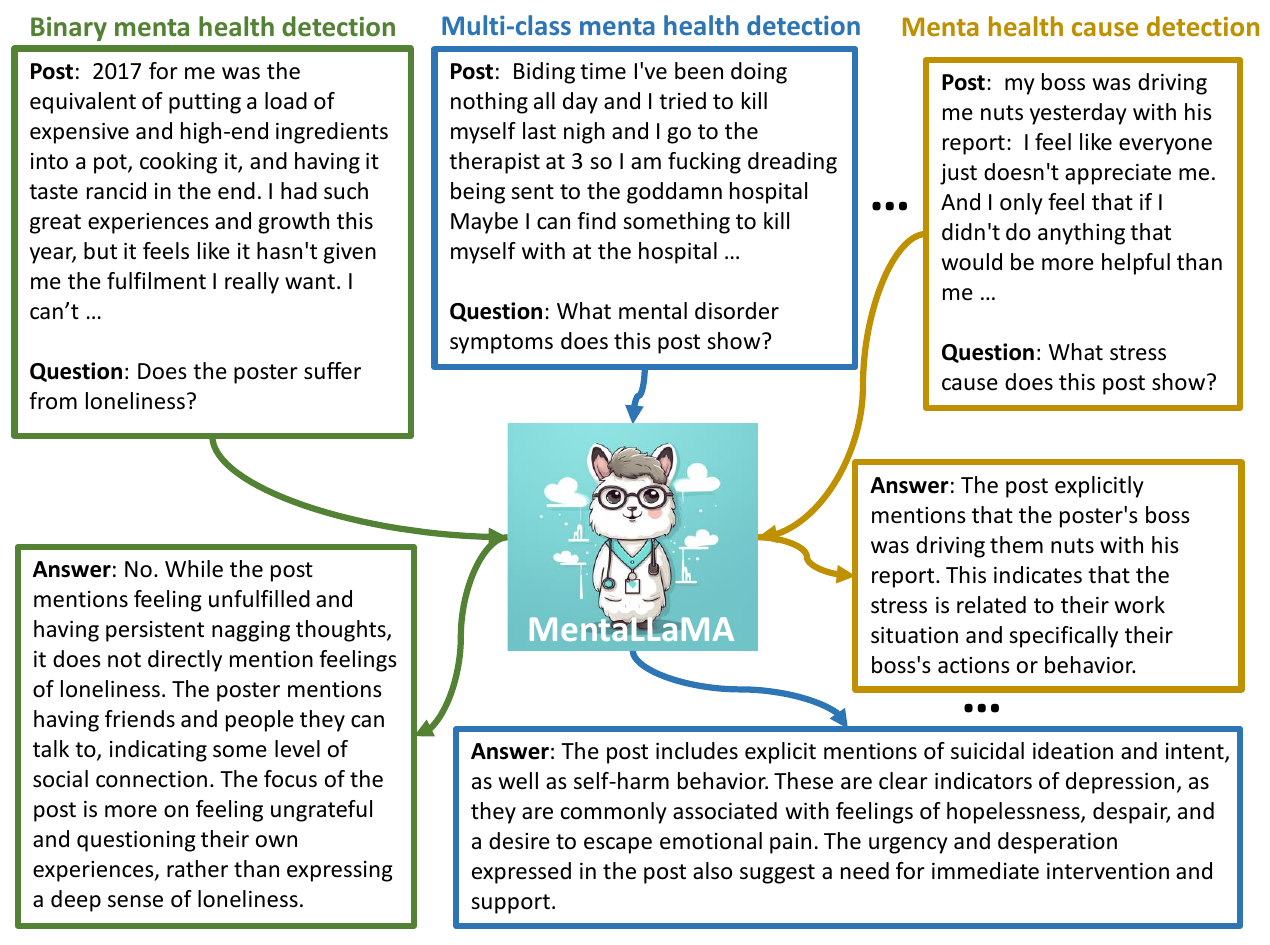}
\caption{Some examples of MentaLLaMA's capabilities in diverse mental health analysis tasks.}
\label{fig:demo}
\end{figure}

Firstly, we collect raw data from 10 existing data sources covering 8 mental health analysis tasks. The collected data includes social media posts and their corresponding annotations for mental health-related tasks. Secondly, inspired by the success of self-instruct~\cite{wang-etal-2023-self-instruct} and ChatGPT's great potential in generating human-level explanations for mental health analysis~\cite{yang2023evaluations}, we use expert-written few-shot examples and collected annotations to prompt ChatGPT to obtain a high-quality explanation for each annotation. 
To ensure the quality of the explanations, we perform comprehensive automatic evaluations on all collected data, where the correctness of the predictions, consistency between annotations and explanations, and quality of the explanations are evaluated. We also perform human evaluations for a subset of the collected data with a carefully designed annotation scheme from domain experts. Thirdly, we transform all collected social media posts, the annotations, and the explanations into instruction-based query-answer pairs in a rule-based manner, which are used to build the IMHI training data and the IMHI evaluation benchmark, the first holistic evaluation benchmark for interpretable mental health analysis tasks. 

Drawing on the IMHI dataset, we propose MentaLLaMA, the first open-source LLM series based on the LLaMA2 foundation models~\cite{touvron2023llama} for interpretable mental health analysis with instruction-following capability. Specifically, we fine-tune 3 MentaLLaMA models with different model sizes: MentaLLaMA-7B, MentaLLaMA-chat-7B, and MentaLLaMA-chat-13B (some examples of MentaLLaMA's strong capabilities are presented in Figure \ref{fig:demo}). We comprehensively evaluate the performance of MentaLLaMA models and other advanced methods on the IMHI evaluation benchmark. Our focus is twofold: the correctness of mental health detection and the quality of generated explanations. The results show that MentaLLaMA-chat-13B surpasses or approaches SOTA discriminative methods~\cite{ji-etal-2022-mentalbert} on 7 out of 10 test sets in the correctness of the prediction, and MentaLLaMA produces explanations on par with ChatGPT, consistently delivering superior results compared to generative PLMs. Its generation quality benefits from instruction tuning, reinforcement learning from human feedback (RLHF)~\cite{stiennon2020learning}, and increasing model sizes. 
MentaLLaMA models also show strong generalizability to unseen tasks, which show better predictive correctness than ChatGPT and outclass generative PLMs in the quality of explanations.

We summarize our contributions as follows: 1) We formalize the interpretable mental health analysis task and build the IMHI dataset, the first multi-task and multi-source instruction-tuning dataset for interpretable mental health analysis on social media. 2) We propose MentaLLaMA, the first open-source instruction-following LLM series for interpretable mental health analysis. 3) We introduce the first holistic evaluation benchmark with 19K test samples, which covers 8 tasks and 10 test sets. 4) Our results and analysis on this benchmark demonstrate the superiority of MentaLLaMA on prediction correctness, the quality of generated explanations, and generalization ability

\begin{table*}[]
\caption{Statistics of the collected data. "Raw" and "Instruction" denote the split sample numbers for the raw data and converted instruction data in the IMHI dataset. "Annotation" denotes the reliability of the annotations in the raw data.}\label{raw_statistic}
\resizebox{.9\textwidth}{!}{
\begin{tabular}{llllllr}
\toprule
\textbf{Data} & \textbf{Task} & \textbf{Raw(train/val/test)} & \textbf{Instruction(train/val/test)} & \textbf{Source} & \textbf{Annotation} & \textbf{Labels/Aspects} \\ \midrule
DR & depression detection & 1,003/430/405&1,003/430/405 & Reddit & weak supervision & Yes, No \\
Dreaddit & stress detection & 2,837/300/414 & 2,837/300/414 & Reddit & human annotation & Yes, No \\
CLP & depression detection & 456/196/299& 456/196/299 & Reddit & human annotation & Yes, No \\
SWMH & mental disorders detection & 34,822/8,705/10,882& 34,822/8,705/10,882 & Reddit & weak supervision & Suicide, Anxiety, Bipolar disorder, Depression, None \\
T-SID & mental disorders detection & 3,071/767/959& 3,071/767/959 & Twitter & weak supervision & None, Suicide, Depression, PTSD \\
SAD & stress cause detection & 5,547/616/684& 5,547/616/684 & SMS & human annotation & \begin{tabular}[c]{@{}r@{}}School, Finance, Family, Social Relation,\\ Work, Health, Emotion, Decision, Others\end{tabular} \\
CAMS & depression/suicide cause detection & 2,207/320/625& 2,207/320/625 & Reddit & human annotation & \begin{tabular}[c]{@{}r@{}}Bias, Jobs, Medication, Relationship, \\ Alienation, None\end{tabular} \\
loneliness & loneliness detection & 2,463/527/531& 2,463/527/531 & Reddit & human annotation & Yes, No \\
MultiWD & Wellness dimensions detection & 2,624/250/353 & 15,744/1,500/2,441 & Reddit & human annotation & \begin{tabular}[c]{@{}r@{}}Spiritual, Physical, Intellectual, Social, \\ Vocational, Emotional\end{tabular} \\
IRF & interpersonal risk factors detection & 1,971/493/1,059 & 3,943/985/2,113 & Reddit & human annotation & Thwarted Belongingness, Perceived Burdensomeness \\ \bottomrule
\end{tabular}}
\end{table*}

\section{Task Formalization}
Based on preliminary explorations~\cite{yang2023evaluations,xu2023leveraging}, we formally define the interpretable mental health analysis task in this section. 
Unlike previous discriminative settings, we model mental health analysis as a generation task, where a generative model, such as an autoregressive language model $P_\phi(y|x)$ parameterized by pre-trained weights $\phi$, is set as the foundation. 
The model is adapted to simultaneously solve $N$ mental health analysis tasks, such as mental health detection and cause detection, and generate explanations for the decisions. 
Each task $t$ is represented by a subset of $N_t$ training context-target pairs: $\mathcal{D}_t=\{(q_i^t, r_i^t)\}_{i=1,...,N_t}$, where $q$ is a token sequence containing the target post and the query, and $r$ is another sequence consisting of the answer to the query (e.g. the classification result) and a rationale for the decision making conveyed in natural language. All subsets are merged as the training dataset: $\mathcal{D}=\mathop{\cup}_{t=1,...,N}\mathcal{D}_t$. The model is optimized on these data to improve the correctness of predictions and the quality of rationales by maximizing the conditional language modeling objective:
\begin{equation}
    \mathop{max}\limits_{\phi}\sum\limits_{(q, r)\in\mathcal{D}}\sum\limits_{j=1}^{|r|}\mathop{log}(P_\phi(r_j|q, r_{<j}))
\end{equation}

\begin{figure}[htpb]
\centering
\includegraphics[width=7cm,height=4.67cm]{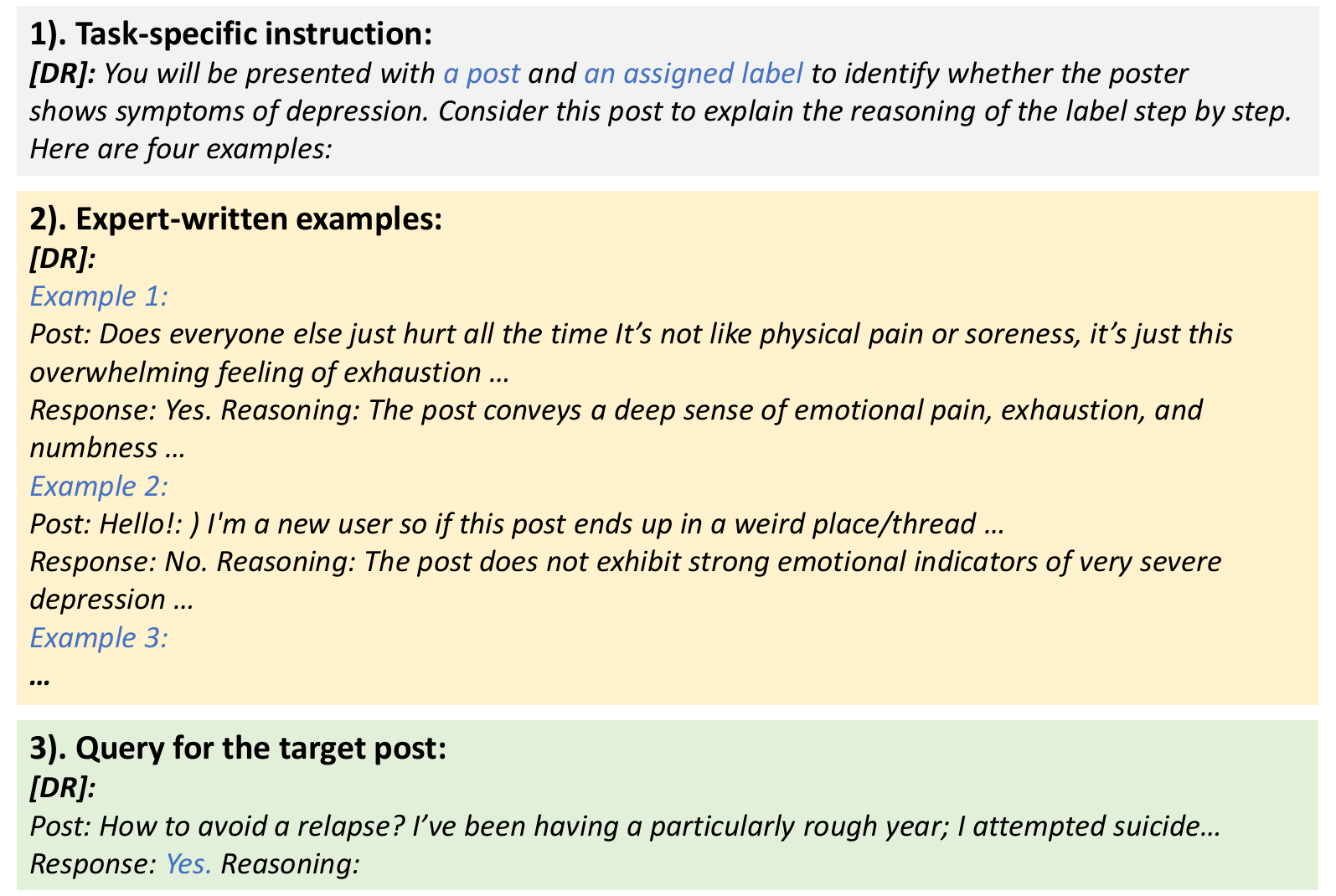}
\caption{Three components are concatenated to construct the prompts. The key information is marked in blue.}
\label{fig:example_prompt}
\end{figure}

\section{IMHI Dataset}
This section introduces the construction process of the IMHI dataset. The process mainly involves 4 procedures: raw data collection, explanation generation via ChatGPT, evaluation for the generated explanations, and instruction construction.

\subsection{Raw Data Collection}
The raw data is collected from 10 existing mental health analysis datasets from multiple social media data sources, including Reddit, Twitter, and Short Message Service (SMS) texts. These datasets are also with high-quality annotations, which are important resources for explanation generation and AIGC evaluation. More statistics of the collected raw data are shown below and in Table \ref{raw_statistic}.

\textbf{Binary mental health detection}.
This task aims to detect symptoms of one mental health condition, where each social media post is annotated with a binary label. We select two datasets for depression symptom detection: Depression\_Reddit (DR)~\cite{pirina2018identifying} and CLPsych15 (CLP)~\cite{coppersmith2015clpsych}. We also utilize Dreaddit~\cite{turcan2019dreaddit}, a dataset for stress detection, and a loneliness symptom detection dataset.

\textbf{Multi-class mental health detection}.
This task aims to identify symptoms of one mental health condition from a given list of multiple mental health conditions, which are normally modeled as a multi-class single-label classification task. We select T-SID~\cite{ji2022suicidal} and SWMH~\cite{ji2022suicidal} datasets for this task, including symptoms of depression, PTSD, anxiety, etc.

\textbf{Mental health cause/factor detection}.
With a post showing a mental health condition, this task aims to assign a label to the post for a possible cause/factor leading to the mental health condition from a given causes/factors list. Common causes include social relationships, medication, work pressure, etc. We select a stress-cause detection dataset SAD~\cite{mauriello2021sad} and a depression/suicide cause detection dataset CAMS~\cite{garg-etal-2022-cams}.

\textbf{Mental risk/wellness factors detection}.
This task dives deep into the social or mental factors behind mental health conditions and aims to identify psychological risk/wellness factors from social media posts, which is also modeled as a classification task to detect the existence of certain factors. We select IRF~\cite{garg-etal-2023-annotated}, an annotated dataset for interpersonal risk factors of mental disturbance. Another dataset called MultiWD~\cite{sathvik2023multiwd} is also collected, which is developed for analyzing mental wellness dimensions from psychological models.

\subsection{Explanation Generation with ChatGPT}\label{instruction_template}

\begin{figure*}[htpb]
\centering
\includegraphics[width=17cm,height=4.25cm]{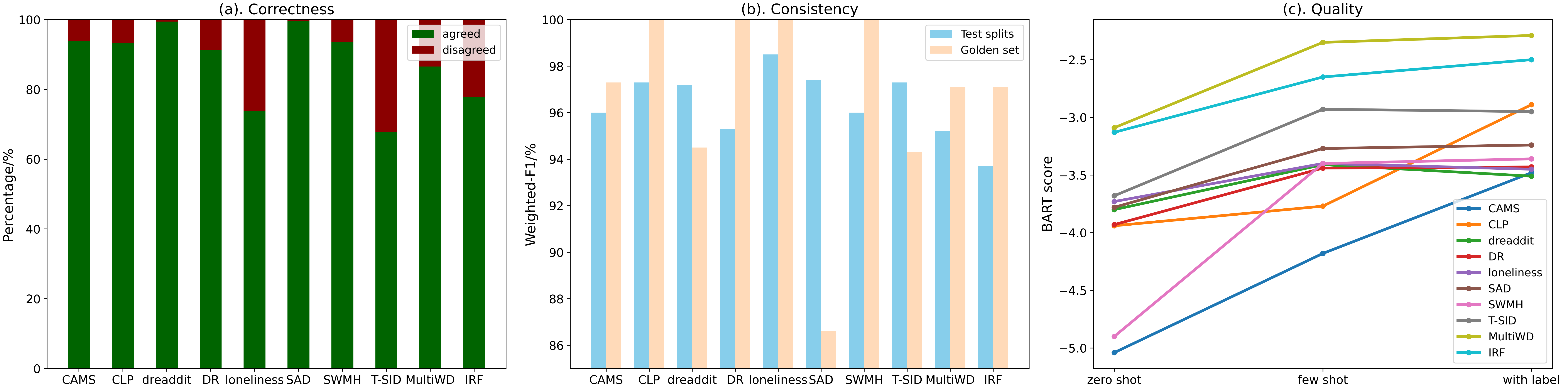}
\caption{Automatic evaluation results on ChatGPT-generated data.}
\label{fig:imhi_evaluation}
\end{figure*}

Though rich data sources with high-quality classification annotations are available, it lacks open-source data that provides detailed and reliable explanations for the annotations. Therefore, we leverage ChatGPT to generate explanations for the collected samples, which is proven a reliable LLM in interpretable mental health analysis~\cite{yang2023evaluations}. Firstly, we ask the domain experts to manually write 1 task-specific instruction and 35 explanation examples for each of the tasks in 10 collected datasets. The expert-written explanations lead to a gold explanation set $\mathcal{G}$ with 350 samples. To facilitate model training and evaluation, all expert-written explanations are based on the following template:
\begin{center}
    \fbox{\textit{[label]}\ \ Reasoning:\ \ \textit{[explanation]}}
\end{center}
where \textit{[label]} and \textit{[explanation]} denote the classification annotation and the corresponding explanation content. Secondly, for each dataset, we randomly sample 2 explanations from $\mathcal{G}$ for each class, and include them as few-shot examples in the prompt. To further enhance the generation quality, we include supervised annotations from the raw datasets. Thirdly, we utilize task-specific instruction, few-shot expert-written examples, and the assigned annotation for the target post to construct the prompt for ChatGPT explanation generation. An example of the constructed prompt for the dataset DR is shown in Figure \ref{fig:example_prompt}, and the prompts for other datasets are presented in Table \ref{prompt_chatgpt} in Appendix.

\subsection{Explanation Evaluation}
We perform comprehensive evaluations on the ChatGPT-generated explanations to ensure their quality. Due to the large quantity of generated explanations (105K), we perform holistic automatic evaluations on all collected data and select a subset for human evaluation.

\subsubsection{Automatic Evaluation}\label{auto_eval}
In automatic evaluation, we believe three criteria are crucial to guarantee the quality of the generated explanations: 1) \textbf{Correctness}: the explanations should make correct label predictions in the corresponding mental health analysis task. 2) \textbf{Consistency}: the explanations should provide clues and analyses that are consistent with their predicted labels~\cite{DBLP:conf/iclr/0002WSLCNCZ23}. 3) \textbf{Quality}: from the perspective of psychology, the generated explanations should provide supportive evidence with high quality in aspects such as reliability, professionality, etc~\cite{yang2023evaluations}.
Based on the above definitions, we design automatic evaluation methods for each of these criteria as follows:

\textbf{Correctness}.
During the explanation generation process, we combine the annotated labels from each collected dataset into the prompts to supervise ChatGPT in generating correct explanations. An appropriate assumption is that a classification result that is agreed upon by both the dataset annotations and ChatGPT can be considered correct. However, we notice that ChatGPT can sometimes express disagreement with the assigned label in its response. Examples of such disagreements are shown in Appendix \ref{disagree_example}. These disagreements are possibly due to the subjectivity of some tasks and the weakly-supervised annotation processes (as shown in Table \ref{raw_statistic}) of some datasets. In these cases, we ask the domain experts to manually check the prompts and responses to modify/rewrite the classification and explanations. We present the agreement percentages between dataset annotations and ChatGPT for each collected dataset in Figure \ref{fig:imhi_evaluation}(a). According to the results, 7 out of 10 datasets have agreement percentages above 90\%, showing the high correctness of most generated responses. T-SID dataset has an agreement percentage below 70\% because it has weakly-supervised labels obtained by the clustering of subreddits in Reddit~\cite{ji2022suicidal}. loneliness and IRF datasets also have percentages below 80\%, as they are built on relatively subjective tasks such as loneliness detection, and interpersonal risk factors identification.

\textbf{Consistency}.
As all ChatGPT generations follow the template specified in Sec. \ref{instruction_template}, consistency evaluates whether the evidence in \textit{[explanation]} supports \textit{[label]} in each response. Specifically, we split \textit{[explanation]} and \textit{[label]} contents via the "Reasoning:" symbol in each response, and use the \textit{[explanation]} and \textit{[label]} pairs from the ChatGPT responses of each raw training split to train a classifier based on MentalBERT~\cite{ji-etal-2022-mentalbert}. For the $i$-th explanation $\textit{[explanation]}_i$ we have:
\begin{equation}
    \textit{[label]}_i^p = MentalBERT(\textit{[explanation]}_i)
\end{equation}
where $\textit{[label]}_i^p$ is then supervised by the $i$-th label $\textit{[label]}_i$. The intuition behind this method is that the training pairs with higher consistency are expected to supervise a more precise classifier for identifying the supported label given the explanation. To evaluate the precision of the trained classifiers, we test them on both the ChatGPT responses for the test split of each raw dataset, and the expert-written gold explanation set $\mathcal{G}$. The classification performance is presented in Figure \ref{fig:imhi_evaluation}(b). According to the results, all classifiers achieve weighted F1 scores of over 93.5\% on the responses for test splits, which shows a highly stable distribution in consistency between ChatGPT-generated explanations and annotated labels. Test results on the gold explanation set show that the classifiers achieve over 94\% on 9 of 10 datasets, with 4 datasets achieving 100\% performance. These results show that the classifiers can identify the correct explanation and label pairs with very high accuracy, which proves the high consistency of the training data (ChatGPT responses on training splits of the raw datasets). However, the performance on SAD is relatively low (86.6\%). A possible reason is that explanations for some labels (e.g. `School' and `Work', `Family' and `Social Relation'), as shown in Table \ref{raw_statistic}, can have similar semantics, which can be difficult to distinguish. With the above evidence, we conclude that ChatGPT-generated explanations have high consistency with the assigned labels.

\textbf{Quality}.
With careful human evaluations, Yang et al.~\cite{yang2023evaluations} show that ChatGPT can generate approaching-human explanations in a zero-shot manner in terms of fluency, reliability, etc. Therefore, we set the zero-shot explanations of ChatGPT as the baseline to evaluate the generation quality of our data. Specifically, based on our designed prompts (we refer to as \textit{with-label prompts}) in Sec. \ref{instruction_template}, we remove the assigned labels to obtain the few-shot prompts, and remove the assigned labels and the few-shot expert-written examples to obtain the zero-shot prompts. We separately use the zero-shot prompts, few-shot prompts, and with-label prompts to probe ChatGPT for the 350 posts in the gold explanation set $\mathcal{G}$. Setting expert-written explanations in $\mathcal{G}$ as the gold standard, we utilize BART-score~\cite{yuan2021bartscore} to automatically evaluate the quality of the responses to the three kinds of prompts, as BART-score is proven most correlated with human evaluations compared to other popular automatic metrics in interpretable mental health analysis~\cite{yang2023evaluations}. The evaluation results are shown in Figure \ref{fig:imhi_evaluation}(c). According to the results, few-shot outputs show significant improvement over zero-shot outputs on all raw datasets, which proves the effectiveness of expert-written few-shot examples in enhancing the quality of the ChatGPT-generated explanations. In addition, the generated explanations from with-label prompts further outperform zero-shot explanations, which are proven to approach human performance. The above evidence proves that the explanations in the IMHI dataset bear high quality.

\begin{figure}[htpb]
\centering
\includegraphics[width=7cm,height=3.5cm]{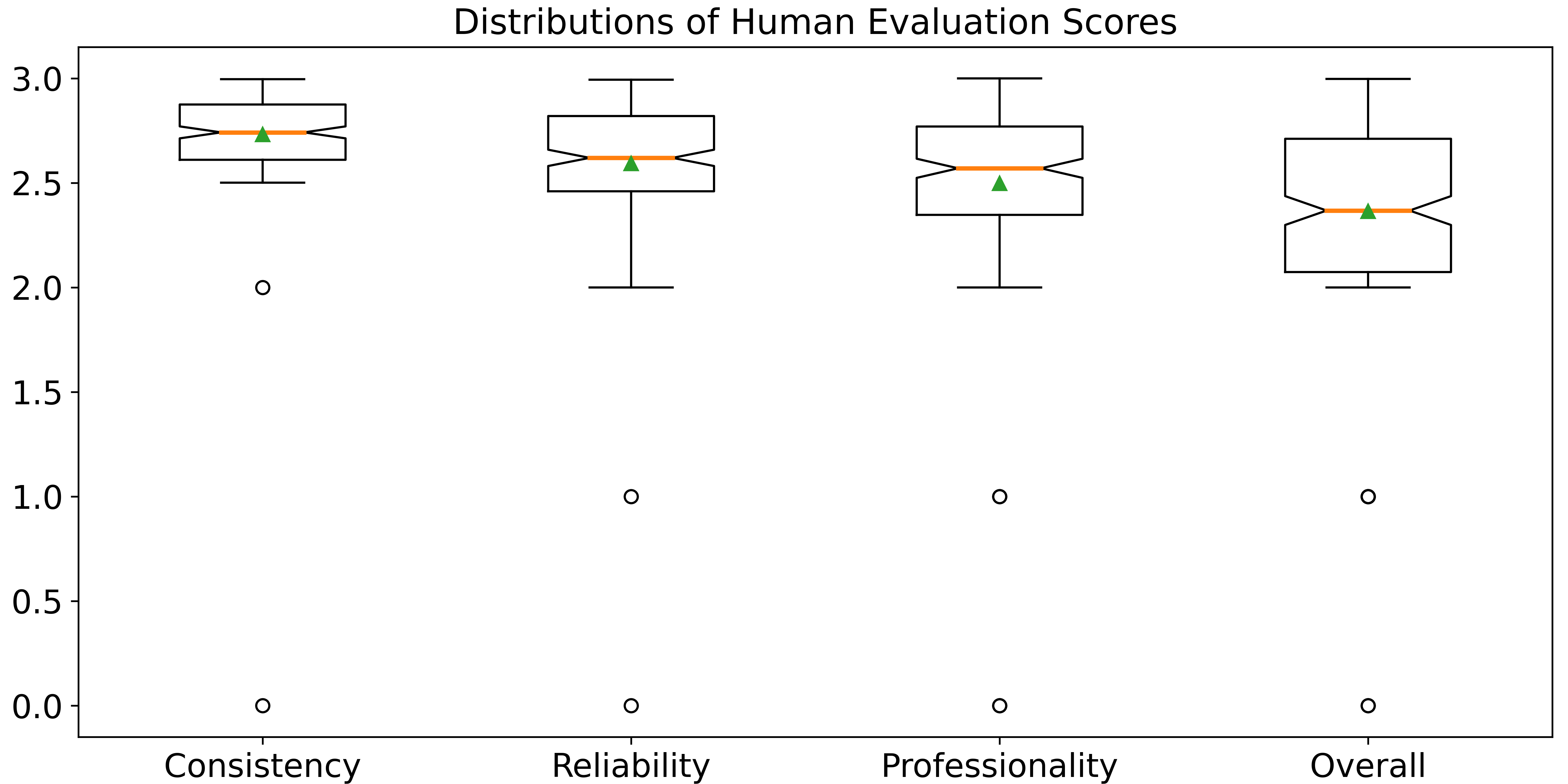}
\caption{Distributions of human evaluation scores on ChatGPT-generated explanations. Orange lines and green dots denote the median and average numbers.}
\label{fig:human_chatgpt}
\end{figure}

\subsubsection{Human Evaluation}\label{human_settings}
We randomly select 200 explanations generated from the raw datasets to perform human evaluations. The annotation scheme is developed based on previous protocols for similar tasks~\cite{yang2023evaluations,wallace2021generating}, and further modified for interpretable mental health analysis with collaborative efforts from 2 domain experts (Quantitative Psychology Ph.D. students). Specifically, we assess the explanations in 4 aspects:
1) \textbf{Consistency}: The text should be built from sentence to sentence to a coherent body of information about mental health that supports the classification results.
2) \textbf{Reliability}: The trustworthiness of the evidence to support the classification results in the generated explanations.
3) \textbf{Professionality}: It measures the rationality of the evidence in generated explanations from the perspective of psychology.
4) \textbf{Overall}: The general effectiveness of the generated explanation.
Each aspect is divided into 4 standards rating from 0 to 3, where higher scores denote more satisfactory performance. More details of the annotation scheme are presented in Appendix \ref{scheme}. During annotation, each sample is rated by 3 domain experts (Quantitative Psychology Ph.D. students) on all aspects. We aggregate all annotations by averaging the scores of each sample and present the results in Figure \ref{fig:human_chatgpt}. According to the results, most explanations are assigned consistency scores over 2.5, which shows that these data are consistent with the classification results, and completely fluent, coherent, and error-free. Most samples also obtain over 2.0 scores on reliability, proving that they provide mostly reliable information with non-critical misinformation or wrong reasoning. Finally, the evaluation results on professionality indicate that most explanations can provide multiple evidences that are supportive from the perspective of psychology. Overall, the human evaluations show that ChatGPT can generate explanations that have good overall performance, which is consistent with previous analysis~\cite{yang2023evaluations} and the automatic evaluation results.

\subsection{Instruction Construction}
We construct the IMHI dataset based on all posts from the raw datasets and the corresponding evaluated ChatGPT-generated explanations. We simplify the instructions introduced in Sec. \ref{instruction_template} to adapt to less powerful LLMs and construct the questions in a rule-based manner. The evaluated ChatGPT-generated explanations are directly used as the responses to these questions. We mix the question-response pairs from the training split of all raw datasets and randomize the order to build the training split of the IMHI dataset, which consists of 72,095 samples. To facilitate the best model selection, we build a validation set, which is developed from the valid split of each raw dataset using the same method, with 14,346 samples.

Due to the poor instruction following ability of some baseline models, we also convert the IMHI data into a completion-based form using another set of templates. We refer to this dataset as IMHI-completion. The prompt templates for IMHI and IMHI-completion datasets are presented in Appendix \ref{IMHI_complete}

\section{MentaLLaMA Training}
Based on the IMHI dataset, we finetune the LLaMA2~\cite{touvron2023llama} models to build our MentaLLaMA models. Firstly, we build a MentaLLaMA-7B by training LLaMA2-7B on the IMHI training set for 10 epochs, and select the best model based on the validation results on the IMHI validation set. We set the batch size to 32 and a gradient accumulation step of 8, which leads to an actual batch size of 256. The model is trained based on the AdamW optimizer~\cite{loshchilov2017decoupled}, and we set a max learning rate of 1e-5 with a warm-up ratio of 3\%. The max model input length is set to 2048. We also utilize Flash-Attention~\cite{dao2022flashattention} to accelerate the training process. Secondly, we build MentaLLaMA-chat-7B and MentaLLaMA-chat-13B models by training on LLaMA2-chat-7B and LLaMA2-chat-13B, which are optimized with instruction tuning~\cite{ouyang2022training}, and the first open-source LLMs tuned with reinforcement learning from human feedback (RLHF)~\cite{stiennon2020learning}. The training process is on the same IMHI dataset with the same experimental settings. Thirdly, to enable fair comparisons with the baseline models that are fine-tuned in a completion-based manner, we train another LLaMA2-7B model on the IMHI-completion dataset. All models are trained on 4 Nvidia Tesla A100 GPUs, each with 80GB of memory.

\section{IMHI Evaluation Benchmark}
We build the IMHI evaluation benchmark for interpretable mental health analysis on the test splits of the collected datasets. As data from each dataset requires a different evaluation metric setting, we split the test data into 10 subsets based on the data sources. The statistics of the evaluation benchmark are presented in Table \ref{raw_statistic}.


Following the evaluation criteria of AIGC introduced in Sec. \ref{auto_eval}, the benchmark evaluates 2 key aspects of the model responses: correctness of the predictions and quality of the explanations. We model the evaluation of correctness as a classification task and compute the weighted F1 scores based on the predictions of the output and the assigned labels in the references. A key challenge of this method is that some models, especially the instruction-tuned ones, do not respond in a unified template as in Sec. \ref{instruction_template}. These irregular responses make rule-based determinations of the predicted labels difficult. To solve this problem, we utilize the MentalBERT-based classifiers, which are used for evaluating the consistency of the IMHI dataset (introduced in Sec. \ref{auto_eval}), to assign a prediction label to each response. The classifiers are expected to accurately assign the labels based on the responses because they are proven to perform well in the IMHI test set and the gold explanation set, as shown in Figure \ref{fig:imhi_evaluation}(b).
For evaluating the explanation quality, we follow the same methods as in Sec. \ref{auto_eval}, where BART-score~\cite{yuan2021bartscore} is used to evaluate the model outputs.

\begin{table*}[!hbt]
\caption{Evaluation results of correctness on the IMHI test set. All results are weighted F1 scores. "Param." denotes the number of parameters for each model. In zero-shot/few-shot Methods, "ZS" denotes zero-shot methods, and "FS" denotes few-shot methods. The best values in discriminative and interpretable mental health analysis methods are highlighted in bold.}\label{tab:IMHI-correct-results}
\resizebox{.8\textwidth}{!}{
\begin{tabular}{lc|cccccccccc}
\toprule
\textbf{Model} & \textbf{Param.} & \textbf{CAMS} & \textbf{CLP} & \textbf{DR} & \textbf{Dreaddit} & \textbf{IRF} & \textbf{loneliness} & \textbf{MultiWD} & \textbf{SAD} & \textbf{SWMH} & \textbf{T-SID}\\
\midrule
\multicolumn{12}{c}{\textbf{Discriminative methods}}\\
BERT-base & 110M & 34.92 & 62.75 & 90.90 & 78.26 & 72.30 & 83.92 & \textbf{76.69} & 62.72 & 70.76 & 88.51 \\
RoBERTa-base & 110M & 36.54 & 66.07 & 95.11 & 80.56 & 71.35 & 83.95 & -- & 67.53 & 72.03 & 88.76\\
MentalBERT & 110M & 39.73 & 62.63 & \textbf{94.62} & 80.04 & \textbf{76.73} & 82.97 & 76.19 & 67.34 & 71.11 & 88.61 \\
MentalRoBERTa & 110M & \textbf{47.62} & \textbf{69.71} & 94.23 & \textbf{81.76} & -- & \textbf{85.33} & -- & \textbf{68.44} & \textbf{72.16} & \textbf{89.01}\\ \midrule
\multicolumn{12}{c}{\textbf{Zero-shot/few-shot methods}}\\
LLaMA2-7B$_{ZS}$ & 7B & 16.34 & 36.26 & 58.91 & 53.51 & 38.02 & 58.32 & 40.1 & 11.04 & 37.33 & 25.55 \\
LLaMA2-13B$_{ZS}$ & 13B & 14.64 & 39.29 & 54.07 & 36.28 & 38.89 & 55.48 & 53.65 & 13.2 & 40.5 & 25.27\\
ChatGPT$_{ZS}$ & 175B & 33.85 & 56.31 & 82.41 & 71.79 & 41.33 & 58.40 & 62.72 & 54.05 & 49.32 & 33.30\\ 
ChatGPT$_{FS}$& 175B & 44.46 & 61.63 & 84.22 & 75.38 & 43.31 & 58.78 & 64.93 & 63.56 & 60.19 & 43.95\\
GPT-4$_{FS}$& 1.76T & 42.37 & \textbf{62.0} & 82.0 & 78.18 & 51.75 & 72.85 & 62.58 & 55.68 & 62.94 & 40.48\\
\multicolumn{12}{c}{\textbf{Completion-based fine-tuning methods}}\\
T5-Large & 770M & 40.2 & 48.6 & 84.9 & 77.7 & 74.0 & 80.8 & 76.4 & 58.1 & 70.0 & 77.1\\
BART-Large & 406M & 43.8 & 50.3 & 84.6 & \textbf{80.0} & 76.2 & 83.3 & \textbf{77.2} & 59.6 & 71.5 & \textbf{77.9}\\
LLaMA2-7B & 7B & 30.47 & 51.17 & 84.94 & 61.59 & 73.5 & 81.25 & 65.52 & 49.6 & 63.08 & 68.93\\
\multicolumn{12}{c}{\textbf{Instruction-tuning methods}}\\
MentaLLaMA-7B & 7B & 32.52 & 59.86 & 76.14 & 71.65 & 67.53 & 83.52 & 68.44 & 49.93 & 72.51 & 72.64\\
MentaLLaMA-chat-7B & 7B & 44.8 & 51.84 & 83.95 & 62.2 & 72.88 & 83.71 & 75.79 & 62.18 & \textbf{75.58} & 77.74\\
MentaLLaMA-chat-13B & 13B & \textbf{45.52} & 52.61 & \textbf{85.68} & 75.79 & \textbf{76.49} & \textbf{85.1} & 75.11 & \textbf{63.62} & 71.7 & 75.31 \\
\bottomrule
\end{tabular}}
\end{table*}

\section{Experiments and Analysis}
\subsection{Baseline Models}
We select the following strong and representative baseline models to compare with our MentaLLaMA models:

\textbf{Discriminative methods}.
As mental health analysis is previously modeled as text classification tasks, we select classification models as baseline models, where most recent methods finetune discriminative PLMs such as BERT~\cite{devlin-etal-2019-bert} and RoBERTa~\cite{liu2019roberta} on the target dataset. We also include SOTA methods MentalBERT and MentalRoBERTa~\cite{ji-etal-2022-mentalbert}, which pre-train a language model from scratch on large-scale data in the mental health domain and further finetune on the target datasets. As all these models cannot generate texts, we only use these models in comparisons of correctness.

\textbf{Zero-shot/few-shot methods}.
With the recent advancement in foundation LLMs, zero-shot and few-shot solutions have become effective and cost-efficient. We select the 7B and 13B versions of the open-source LLM LLaMA2~\cite{touvron2023llama} to perform zero-shot prompting on the benchmark data. We also perform zero-shot and few-shot prompting on the close-source LLM ChatGPT and GPT-4~\cite{OpenAI2023GPT4TR}.

\textbf{Completion-based fine-tuning methods}.
To evaluate the parameter efficiency of our models, we also finetune generative PLMs with smaller sizes with the same training settings. We select SOTA generative PLMs BART-large~\cite{lewis-etal-2020-bart} and T5-large~\cite{raffel2020exploring}. Since these PLMs do not possess strong instruction-following ability~\cite{ouyang2022training}, we finetune them on the IMHI-completion dataset. To enable fair comparison, we also train a LLaMA-7B model on the same dataset.


\subsection{IMHI Test Results}
\subsubsection{Correctness}
The evaluation results of correctness are presented in Table \ref{tab:IMHI-correct-results}. In discriminative methods, MentalBERT and MentalRoBERTa still achieve SOTA performance on 8 out of 10 test sets. Considering the small sizes of these models, we conclude that fine-tuning domain-specific PLMs remains the most efficient method for discriminative mental health analysis. However, the key limitation of these methods is the poor generalization ability and interpretability of their decisions. Their ability is limited to the originally trained tasks and it's hard to understand how they make decisions. In comparisons between zero-shot methods, ChatGPT$_{ZS}$ significantly outperforms both LLaMA2 models on all 10 datasets. These results are possibly due to the emergent ability~\cite{wei2022emergent} of LLMs, where the mental health analysis ability is weak in smaller models (7B, 13B LLaMA2 models), but rapidly improves in larger models (175B ChatGPT). In addition, ChatGPT$_{FS}$ and GPT-4$_{FS}$ further outperforms ChatGPT$_{ZS}$ on all test sets. These observations are consistent with previous works~\cite{brown2020language}, where in-context learning from expert-written examples can calibrate LLMs' decision boundaries for subjective tasks. However, GPT-4 does not show apparent advantages over ChatGPT on most datasets. All fine-tuning methods show significant improvement over LLaMA2$_{ZS}$ results on all datasets, which generally proves the effectiveness of completion/instruction-based fine-tuning. In completion-based fine-tuning methods, we surprisingly find that T5 or BART outperforms LLaMA2-7B on most test sets with only 15\% in model size. A possible reason is that training LLaMA2 on the unnatural IMHI-completion dataset cannot trigger its ability well. To further evaluate this hypothesis, we train MentaLLaMA-7B with the IMHI dataset. As shown, MentaLLaMA-7B outperforms the completion-based LLaMA2-7B on 8 out of 10 test sets, showing domain-specific instruction tuning as more efficient than completion-based finetuning in improving the correctness of LLaMA2. Experiments on LLaMA2-chat further prove this conclusion, as MentaLLaMA-chat-7B and MentaLLaMA-chat-13B outperform MentaLLaMA-7B on 9 out of 10 test sets. Based on LLaMA2, LLaMA2-chat models are enhanced with high-quality instruction tuning~\cite{ouyang2022training}, which allows them to better follow the mental health-related questions. Notably, MentaLLaMA-chat-13B surpasses or bears a less than 5\% gap to MentalRoBERTa in 7 out of 10 test sets, showing its approaching SOTA ability in achieving correctness in mental health analysis.

\begin{figure*}[htpb]
\centering
\includegraphics[width=15cm,height=7.5cm]{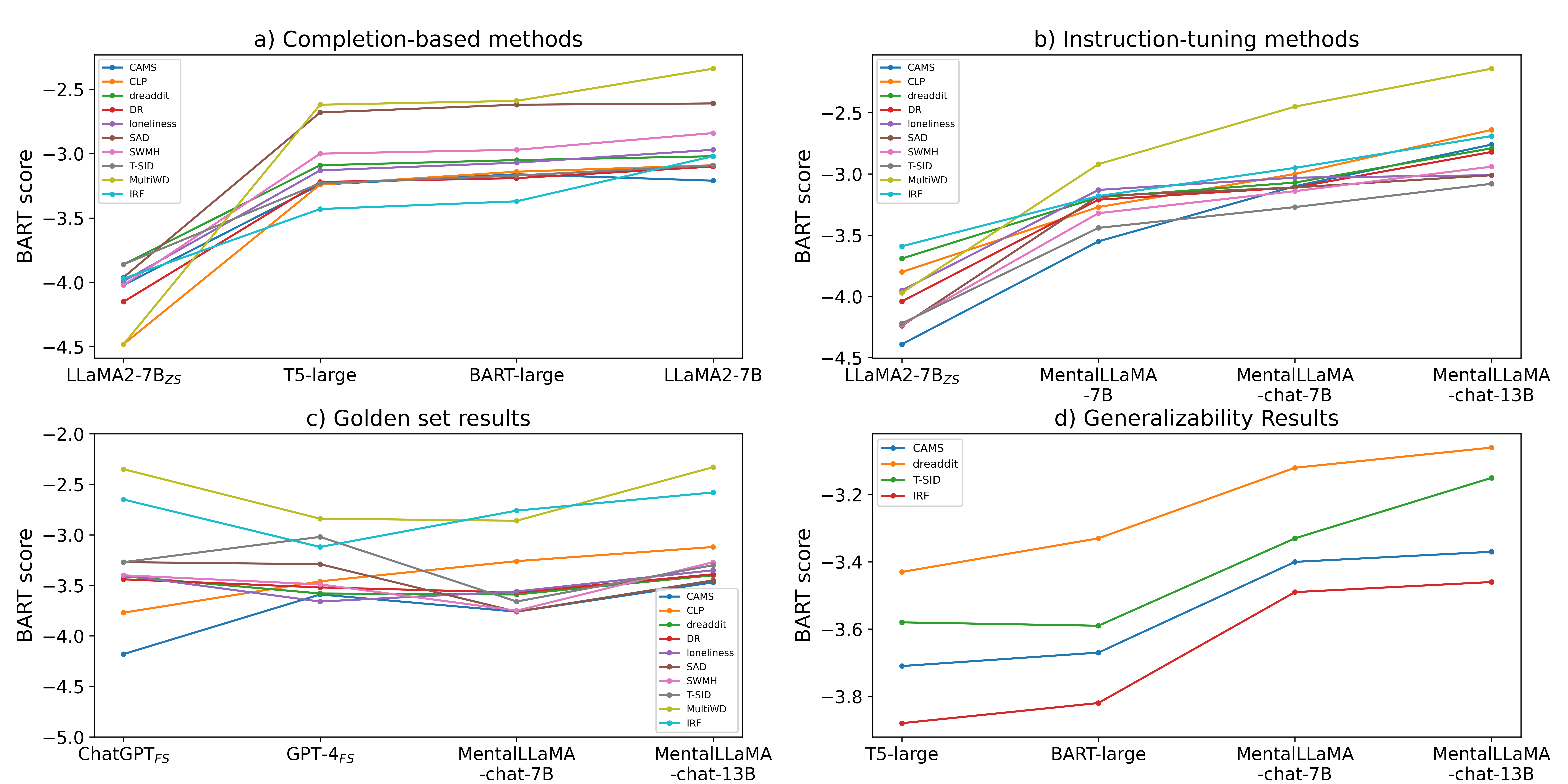}
\caption{BART-score evaluation results on the IMHI test set and expert-written gold set.}
\label{fig:imhi_quality}
\end{figure*}

\subsubsection{Quality}
We present the BART-score evaluation results to evaluate the quality of the explanation generation. In completion-based methods presented in Figure \ref{fig:imhi_quality}(a), LLaMA2-7B greatly outperforms LLaMA2-7B$_{ZS}$ on all 10 test sets, showing the effectiveness of completion-based finetuning in improving the quality of the explanations. T5 and BART models generate explanations that have similar scores, showing their close ability in interpretable text generation. LLaMA2-7B outperforms BART-large on 9 out of 10 test sets, but to a limited scale, where only 2 test sets (MultiWD and IRF) improve over 0.2 in BART-score. These results further prove that completion-based finetuning for LLaMA2 is inefficient. Based on the above observations, we recommend utilizing BART-large to build a completion-based interpretable mental health analysis model, which is both capable and cost-efficient.

In instruction tuning methods presented in Figure \ref{fig:imhi_quality}(b), MentaLLaMA greatly outperforms zero-shot results on LLaMA2-7B on all 10 test sets, showing the effectiveness of instruction tuning in improving the quality of the explanations. MentaLLaMA-chat-7B also significantly outperforms MentaLLaMA-7B, with improvement in all 10 test sets and over 0.2 gain on 6 test sets. These results prove that the instruction tuning and RLHF~\cite{stiennon2020learning} enhancements on LLaMA2-chat models also improve their ability to generate high-quality explanations compared to the vanilla LLaMA2 models. In addition, MentaLLaMA-chat-13B further advances the quality of the explanations, which outperforms MentaLLaMA-chat-7B by over 0.2 on 8 out of 10 test sets. These results show that LLaMA-chat can efficiently leverage the expansion of model size to enhance its interpretability. We believe the RLHF training allows larger models to use their increasing capabilities to generate explanations that are more aligned with human preferences.

We also compare the generation quality of MentaLLaMA on the expert-written gold set $\mathcal{G}$ to the few-shot results on ChatGPT and GPT-4. According to the results in Figure \ref{fig:imhi_quality}(c), the MentaLLaMA models achieve comparable performance to ChatGPT and GPT-4 on most test sets with much smaller model sizes, showing the effectiveness of IMHI instruction tuning and the outstanding explanation generation quality of MentaLLaMA models. We also notice that GPT-4 does not show significant improvement in generation quality over ChatGPT. ChatGPT has comparable model performance in correctness and quality to GPT-4 but with much lower inference costs, which is more appropriate for obtaining large-scale responses for building the IMHI dataset. 

\begin{table}[!hbt]
\caption{Correctness evaluation results on generalizability.}\label{tab:general-correct-results}
\resizebox{.4\textwidth}{!}{
\begin{tabular}{l|cccc}
\toprule
\textbf{Model} & \textbf{CAMS} & \textbf{Dreaddit} & \textbf{IRF} & \textbf{T-SID}\\
\midrule
LLaMA2-13B$_{ZS}$ & 14.64 & 36.28 & 38.89 & 25.27 \\
ChatGPT$_{ZS}$ & \textbf{33.85} & 71.79 & 41.33 & 33.30\\
MentaLLaMA-chat-7B & 20.19 & 67.42 & 54.6 & 64.76 \\
MentaLLaMA-chat-13B & 27.22 & \textbf{71.98} & \textbf{65.51} & \textbf{70.7} \\
\bottomrule
\end{tabular}}
\end{table}

\subsection{Generalizability}
In addition to their outstanding generation ability, LLMs are also proven to bear high generalizability to unseen tasks~\cite{brown2020language,kojima2022large}. To evaluate the generalizability of MentaLLaMA, we exclude the data of the following tasks from the IMHI training set: stress detection (Dreaddit), mental disorder detection from Twitter (T-SID), depression/suicide cause detection (CAMS), and interpersonal risk factors detection (IRF), to build a new training set IMHI-general. We re-finetune T5, BART, and MentaLLaMA-chat models on IMHI-general, and evaluate these models on the test sets of the 4 unseen tasks.

We first evaluate model performance on correctness, where the results are presented in Table \ref{tab:general-correct-results}. As shown, MentaLLaMA models significantly outperform LLaMA2-13B$_{ZS}$ on all datasets, showing the effectiveness of the IMHI instruction tuning in enhancing generalizability to unseen mental health analysis tasks. MentaLLaMA models also outperform ChatGPT$_{ZS}$ on 3 datasets, which further proves their competitive ability in generalizing to the mental health domain.

In terms of explanation quality, the BART-score test results are shown in Figure \ref{fig:imhi_quality}(d). According to the results, MentaLLaMA-chat models significantly outperform T5 and BART on Dreaddit and CAMS, showing that MentaLLaMA-chat models can generate explanations with higher quality to new tasks in fundamental mental health conditions/cause detection tasks. MentaLLaMA's superior performance on IRF also proves its deeper understanding of high-level mental health factors behind mental health conditions. Excluding all Twitter data from the training set, MentaLLaMA-chat models still achieve better scores on the Twitter-derived test set T-SID, proving that MentaLLaMA can be better generalized to new data sources with different data characteristics. In addition, MentaLLaMA-chat-13B further improves the explanation quality compared to MentaLLaMA-chat-7B, denoting the benefit of model size expansion to interpretable mental health analysis on new tasks. Overall, the aforementioned analysis proves that MentaLLaMA bears higher generalizability in unseen tasks compared to other generative PLMs.

\begin{figure}[htpb]
\centering
\includegraphics[width=7cm,height=3.5cm]{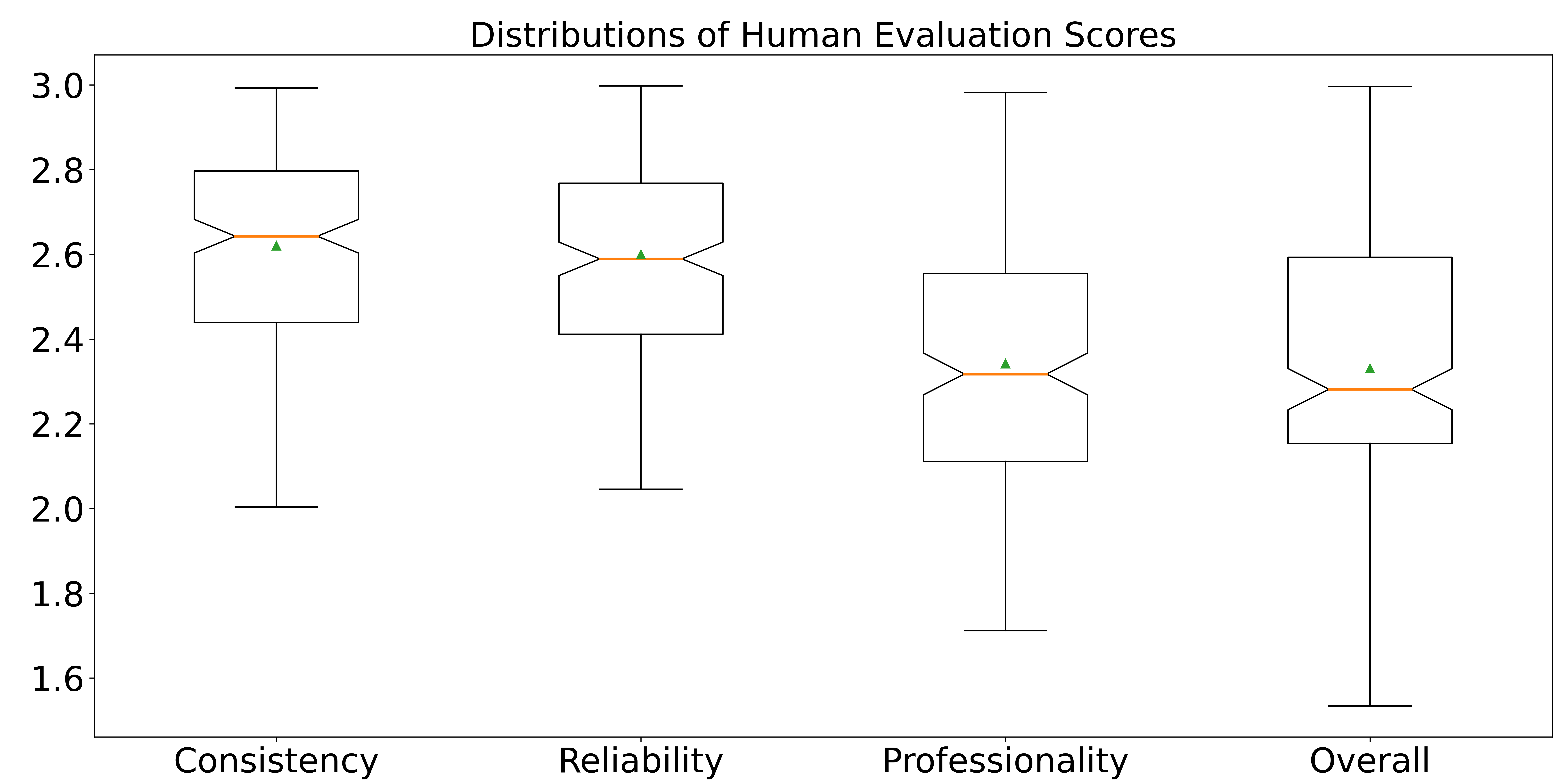}
\caption{Distributions of human evaluation scores on MentaLLaMA-generated explanations.}
\label{fig:mentallama_human_evaluation}
\end{figure}

\subsection{Human Evaluation}
Though automatic evaluations indicate the high quality of the explanations, BART-score is proven to only bear moderate correlations to human evaluation results for interpretable mental health analysis~\cite{yang2023evaluations}, which limits the reliability of its results. Therefore, we further perform human evaluations on 200 random samples from the outputs of MentaLLaMA-chat-13B with the same settings as in Sec. \ref{human_settings}.

As shown in Figure \ref{fig:mentallama_human_evaluation}, MentaLLaMA-generated explanations show general high quality by achieving over 2.2 average scores on all four aspects. Compared to the evaluations of ChatGPT in Figure \ref{fig:human_chatgpt}, MentaLLaMA shows comparable quality in consistency and reliability, proving that the explanations convey a coherent body of information about mental health-related rationales and show high-level trustworthiness to support the prediction results. However, MentalLLaMA significantly underperforms ChatGPT in professionality with a much lower average score. This result shows that MentalLLaMA still lacks domain-specific knowledge compared to ChatGPT. An effective solution could be continual pre-training on high-quality mental health-related data~\cite{han2023medalpaca,xie2023pixiu}, such as textbooks and questionnaires in Psychology.

\section{Conclusion and Future Work}
This paper proposes the novel task of interpretable mental health analysis and the first multi-task and multi-source IMHI dataset with 105K data for instruction tuning.
We leverage ChatGPT to build the training data and perform strict automatic and human evaluations to ensure their reliability. Based on the IMHI dataset, we propose MentaLLaMA, the first open-source LLM series for interpretable mental health analysis with instruction-following capability. Evaluations on the IMHI benchmark show that MentaLLaMA approaches SOTA discriminative methods in correctness and generates human-level explanations. MentaLLaMA also shows high generalizability to unseen tasks.

During experiments, we noticed that MentaLLaMA still bears limitations of lacking domain-specific knowledge compared to powerful models such as ChatGPT. In future work, we will explore continual pre-training of MentaLLaMA on large-scale high-quality mental health-related data to enhance the professionality of its explanations. In addition, BART-score, the automatic evaluation metric used in this work, is proven to only bear moderate correlations
to human evaluation results for interpretable mental health analysis, which limits the reliability of our conclusions. In future work, we will explore customizing more reliable automatic evaluation metrics for interpretable mental health analysis. 

\section{Ethical Considerations}
The raw datasets collected to build our IMHI dataset are from public social media platforms. We strictly follow the privacy protocols~\cite{nicholas2020ethics} and ethical principles~\cite{benton2017ethical} to protect user privacy and guarantee that anonymity is properly applied in all mental health-related texts. In addition, to minimize misuse, all examples provided in our paper are paraphrased and obfuscated utilizing the moderate disguising scheme~\cite{bruckman2002studying}. 

Although experiments on MentaLLaMA show promising performance, we stress that all predicted results and generated explanations should only be used for non-clinical research. Help-seekers should ask for help from professional psychiatrists or clinical practitioners. In addition, recent studies have indicated LLMs can introduce potential bias, such as gender gaps~\cite{ghosh2023chatgpt}. Meanwhile, incorrect prediction results, inappropriate explanations, and over-generalization also illustrate the potential risks of current LLMs. Therefore, there are still many challenges in applying LLMs to real-scenario mental health monitoring systems.

\begin{acks}
This work is supported by the computational shared facility and President’s Doctoral Scholar award, The University of Manchester. This work is supported by the project JPNP20006 from New Energy and Industrial Technology Development Organization
(NEDO), and Artificial Intelligence Research Center, National Institute of Advanced Industrial Science and Technology, Japan.
\end{acks}

\bibliographystyle{ACM-Reference-Format}
\bibliography{reference}

\appendix

\section{Related Work}
\subsection{Mental Health Analysis on Social Media}
In mental health analysis, traditional methods mostly make predictions in a discriminative manner. Effective methods mostly finetune pre-trained language models (PLMs), such as BERT~\cite{devlin-etal-2019-bert} and RoBERTa~\cite{liu2019roberta}, on a small target set~\cite{DBLP:conf/ijcai/ZhangCWZ22,ji2022towards} usually for one mental health condition. To further enhance the PLM representations, some works pre-train language models from scratch with large-scale mental health-related social media data, which usually produce better post representations than general PLMs. Representative works include MentalBERT~\cite{ji-etal-2022-mentalbert}, MentalXLNet~\cite{ji2023domain}, etc.

Though the above black-box models achieve impressive classification performance, there are works exploring interpretable mental health analysis. Some works incorporate metaphor concept mappings as extra features to provide clues on model decisions~\cite{han-etal-2022-hierarchical}. Other works introduced PHQ-9 questionnaire information to assist the predictions~\cite{nguyen-etal-2022-improving,zhang2023phq}. Commonsense knowledge graphs were also leveraged to increase the transparency of PLMs~\cite{harrigian2020models,yang2022mental}. The recent advancements in LLMs take a leap forward for interpretable mental health analysis. Some works~\cite{amin2023will,xu2023leveraging,yang2023evaluations} comprehensively evaluated the performance of general foundation LLMs on various mental health analysis tasks. Xu et al.~\cite{xu2023leveraging} glimpsed the explanation generation ability of LLMs, and Yang et al.~\cite{yang2022mental} holistically evaluated ChatGPT's explanation generation ability with careful human evaluation.

\subsection{Open-source Large Language Models}
Though LLMs such as ChatGPT and GPT-4~\cite{OpenAI2023GPT4TR} achieve general outstanding performance, their closed-source availability affects the development of the research community. Therefore, many efforts have been made to democratize LLMs, such as the LLaMA series~\cite{touvron2023llama} developed by Meta AI. Based on LLaMA, many works tried to replicate ChatGPT-like instruction-following ability by training on large-scale instruction-tuning data~\cite{ouyang2022training}. Representative general instruction-following LLMs include the Alpaca\footnote{\url{https://crfm.stanford.edu/2023/03/13/alpaca.html}} and the Vicuna\footnote{\url{https://lmsys.org/blog/2023-03-30-vicuna/}} model series. Domain-specific instruction tuning also improves LLM performance in certain domains, such as the MedAlpaca~\cite{han2023medalpaca} in the biomedical domain and the Pixiu models~\cite{xie2023pixiu} in the finance domain. In addition, the LLaMA-chat models~\cite{touvron2023llama} are the first open-source LLMs enhanced with reinforcement learning from human feedback (RLHF)~\cite{stiennon2020learning}, which significantly aligns model responses with human preferences.

\begin{table*}[]
\caption{Templates for constructing prompts for IMHI dataset. \textit{[Post]} denotes the target post. \textit{[Aspect]} denotes the detection aspects (shown in Table \ref{raw_statistic}).}\label{imhi_template}
\resizebox{.9\textwidth}{!}{
\begin{tabular}{lll}
\toprule
\textbf{Dataset} & \textbf{IMHI Template} & \textbf{IMHI-complete Template} \\ \midrule
DR & Post: \textit{[Post]} Question: Does the poster suffer from depression? & Post: \textit{[Post]} Question: The answer to the question "Does the poster suffer from depression?" is\\\midrule
Dreaddit & Post: \textit{[Post]} Question: Does the poster suffer from stress? & Post: \textit{[Post]} Question: the answer to the question "Does the poster suffer from stress?" is\\\midrule
CLP & Post: \textit{[Post]} Question: Does the poster suffer from depression?& Post: \textit{[Post]} Question: the answer to the question "Does the poster suffer from depression?" is\\\midrule
SWMH & \begin{tabular}[c]{@{}l@{}} Post: \textit{[Post]} Question: What mental disorder symptoms does\\ this post show?\end{tabular} & Post: \textit{[Post]} Question: this post shows the mental disorder symptoms of\\\midrule
T-SID & Post: \textit{[Post]} Question: What mental disorder does this post show? & Post: \textit{[Post]} Question: this post shows the mental disorder symptoms of\\\midrule
SAD & Post: \textit{[Post]} Question: What stress cause does this post show? & Post: \textit{[Post]} Question: this post shows the stress cause of\\\midrule
CAMS & \begin{tabular}[c]{@{}l@{}} Post: \textit{[Post]} Question: What mental disorder cause does\\ this post show?\end{tabular} & Post: \textit{[Post]} Question: this post shows the mental disorder cause of\\\midrule
loneliness & Post: \textit{[Post]} Question: Does the poster suffer from loneliness? & Post: \textit{[Post]} the answer to the question "Does the poster suffer from loneliness?" is\\\midrule
MultiWD & \begin{tabular}[c]{@{}l@{}}Post: \textit{[Post]} Question: Does the \textit{[Aspect]} mental wellness dimension\\ exist in the post?\end{tabular} & \begin{tabular}[c]{@{}l@{}} Post: \textit{[Post]} Question: the answer to the question "Does the \textit{[Aspect]} mental wellness\\ dimension exist in the post?" is\end{tabular}\\\midrule
IRF & Post: \textit{[Post]} Question: Does the post show risks of \textit{[Aspect]}? & Post: \textit{[Post]} Question: the answer to the question "Does the post show risks of \textit{[Aspect]}?" is\\ \bottomrule
\end{tabular}}
\end{table*}

\section{Human Annotation Scheme}\label{scheme}
Annotators will be given generated explanations from ChatGPT and the expert-written explanations as the correct reference. Annotators will need to score and annotate the generated explanations from the following aspects:

\textbf{Consistency.} Whether the text builds from sentence to sentence to a coherent body of information about mental health that supports the classification results. Annotators should assess if the generated explanation gives consistent supporting evidence to its classifications and is well-structured.
\begin{itemize}
\item 0: Inconsistent with the classification results.
\item 1: Consistent with the classification results, but with poor readability and several errors.
\item 2: Consistent with the classification results. Mostly coherent and easy to read, with few minor errors.
\item 3: Consistent with the classification results. Completely fluent, coherent, and error-free.
\end{itemize}

\textbf{Reliability.} Reliability measures the trustworthiness of the generated explanations to support the classification results. Annotators should assess whether the explanation is based on facts, has misinformation, and wrong reasoning according to the given post.
\begin{itemize}
\item 0: Completely unreliable information with factual hallucination (e.g. non-existent symptoms).
\item 1: Partly reliable information with wrong reasoning based on facts.
\item 2: Mostly reliable information with non-critical misinformation or wrong reasoning.
\item 3: Completely reliable information.
\end{itemize}

\textbf{Professionality.} Professionality measures the rationality of the generated explanations by evaluating the evidence that supports the classification results from the perspective of psychology. Annotators should assess whether the explanation includes the following specified common diagnosis criteria of depression. To ensure the quality of the annotation scheme, we invite our domain experts to develop a list of common symptoms related to depression and sort these symptoms by criticality. The domain experts consult the Patient Health Questionnaire (PHQ-9)\footnote{\url{https://www.apa.org/depression-guideline/patient-health-questionnaire.pdf}} on determining the symptoms and sort these symptoms on their own knowledge.

Specifically, the following symptoms are checked (sorted by criticality):

\begin{itemize}
    \item Suicide ideation: Thoughts that you would be better off dead.
    \item Self-harm ideation: Thoughts of hurting yourself in some way.
    \item Feeling down, depressed, or hopeless.
    \item Self-guilt ideation: Feeling bad about yourself — or that you are a failure or have let yourself or your family down.
    \item \textbf{Symptoms above are classified as with high criticality, and symptoms below are classified as with low criticality.}
    \item Feeling tired or having little energy. Little interest or pleasure in doing things.
    \item Poor appetite or overeating.
    \item Trouble falling or staying asleep, or sleeping too much.
    \item Trouble concentrating on things, such as reading the newspaper or watching television.
    \item Moving or speaking so slowly that other people could have noticed. Or the opposite — being so fidgety or restless that you have been moving around a lot more than usual 
    \item Uncontrollable sexual desire or sexual frigidity.
    \item Other symptoms.
\end{itemize}

Based on the above symptoms, the annotators score the professionality of each explanation with the following criteria:

\begin{itemize}
\item 0: The explanation provides no supportive evidence, or symptoms with high criticality are missing in the explanation.
\item 1: The explanation provides a few supportive evidence, while some symptoms with higher criticality (than provided evidence) are missing.
\item 2: The explanation provides several supportive evidence, while some symptoms with lower criticality (than provided evidence) are missing.
\item 3: The explanation provides all related supportive evidence in the post.
\end{itemize}

\textbf{Overall Score.}  Overall performance measures the general effectiveness of the generated explanation, taking into account the combined scores for consistency, reliability, and professionality.
\begin{itemize}
\item 0: Poor overall performance. Consistency or reliability is fair or below ($<=1$).
\item 1: Fair overall performance. Consistency and reliability are good ($>=2$), and professionality is fair or below ($<=1$).
\item 2: Good overall performance. Consistency and reliability are excellent (=3), but professionality is below excellent ($<=2$)
\item 3: Excellent overall performance. Consistency, reliability, and professionality are all excellent (=3)
\end{itemize}

\section{Examples}
\subsection{Templates for IMHI Dataset}\label{IMHI_complete}
Templates for constructing prompts for the IMHI and IMHI-completion dataset are presented in Table \ref{imhi_template}.


\subsection{Disagreement Examples of ChatGPT}\label{disagree_example}
We provide some examples of ChatGPT's disagreement on the annotated labels in Table \ref{dis_exa}.

\begin{table}[]
\caption{Some examples of ChatGPT's disagreement on the annotated labels.}\label{dis_exa}
\resizebox{.48\textwidth}{!}{
\begin{tabular}{l}
\toprule
\textbf{Examples} \\ \midrule
\begin{tabular}[c]{@{}l@{}} \textit{[CAMS prompt]} \textbf{Post:} My boyfriend, a couple of his friends and a couple of my friends went out tonight\\ to a bar. We were all kind of drunk and my boyfriend's friend asked a friend of mine if he take a shot from\\ her belly button and she said no. He then asked me and u felt bad for him...\\

\textbf{Response:} Relationship. Reasoning: \textcolor{red}{Medication. Reasoning:} The post mentions the poster's anger \\and emotional problems, as well as a recent incident where they engaged in behavior while drunk that\\ upset their boyfriend. The poster also mentions feeling unsure about what's wrong with them.\\ Given their emotional distress and uncertainty...
 \end{tabular}\\\midrule

\begin{tabular}[c]{@{}l@{}} \textit{[T-SID prompt]} \textbf{Post:} My answer on @tGTZ to: Could i be pregnant and how soon before a missed\\ period can i test. pregnancy\\

\textbf{Response:} PTSD. Reasoning: \textcolor{red}{the assigned label is wrong. The label should be No mental disorders.}\\ Reasoning: The content of the post simply discusses pregnancy and testing for it. There is no\\ indication of any distress or symptoms of a mental disorder.\end{tabular}\\\midrule

\begin{tabular}[c]{@{}l@{}} \textit{[loneliness prompt]} \textbf{Post:} Today I was sulking around because I really wanted ramen but if course,\\ since it's New Year's Day, nothing is open. I was already hangry, had made myself some chicken\\ in the oven, but set the heat too low, so it came out frozen. I was fumed …\\

\textbf{Response:} No. Reasoning: \textcolor{red}{Yes.} The post mentions the poster feeling sulky, hungry,\\ and wanting ramen, but being unable to get it because of it being\\ New Year's Day and nothing being open...\end{tabular}\\\midrule

\begin{tabular}[c]{@{}l@{}} \textit{[MultiWD prompt]} \textbf{Post:} i met this girl at walmart she was the cashier she was kinda smiling\\ a little bit and i felt all warm inside why cant i feel like that all the time now i’m back alone in my\\ house i need a girlfriend.\\

\textbf{Response:} Yes. Reasoning: \textcolor{red}{No, the post does not reflect the wellness dimension of the spiritual aspect.}\\ In the post, the individual expresses a desire for companionship and a girlfriend.\\ However, there is no mention of a search for meaning or purpose in human existence...\end{tabular}\\
\bottomrule
\end{tabular}}
\end{table}

\begin{table*}[]
\caption{Examples of the prompts for ChatGPT.}\label{prompt_chatgpt}
\resizebox{1.\textwidth}{!}{
\begin{tabular}{ll}
\toprule
\textbf{Dataset} & \textbf{Example} \\ \midrule
Dreaddit & \begin{tabular}[c]{@{}l@{}} You will be presented with a post and an assigned label to identify whether the poster suffers from stress. Consider the emotions expressed from this post\\ to explain the reasoning of the label step by step. Here are some examples:\\

Post: Its like that, if you want or not. ME: I have no problem, if it takes longer. But you asked my friend for help and let him wait for one hour...\\
Response: No. Reasoning: The poster is expressing frustration and disappointment with someone who asked for help but did not properly prepare for it...\\

Post:...\\

Post: Like, I don't think they happen if I distract myself with the Internet or doing some hobby that I like.  I remember feeling like this at least once a week...\\
Response: No. Reasoning:
\end{tabular}\\\midrule
CLP & \begin{tabular}[c]{@{}l@{}} You will be presented with some posts, and an assigned label to identify whether the poster suffers from depression. Consider the\\ emotions expressed from this post to explain the reasoning of the label step by step. Here are two examples:\\

Post: RT : so gutted the arctic monkeys are headlining T when I can't go had flower of Scotland stuck in my head for three days RT : Going to the...\\
Response: No. Reasoning: Some sentences express negative emotions such as feeling gutted and missing out on events or feeling fat and unhappy...\\

Post:...\\

Post: I just entered to win a NBC themed Soap Box from Fortune Cookie Soap! RT : Keep your friends close and your closer. three days in...\\
Response: No. Reasoning:
\end{tabular}\\\midrule
SWMH & \begin{tabular}[c]{@{}l@{}} You will be presented with a post that possibly shows mental disorders, and an assigned label to show the type of the mental disorder from the following causes list\\: No mental disorders, Suicide, Depression, Anxiety, Bipolar disorder. You must explain the reasoning of the assigned label step by step. Here are some examples:\\

Post: Suicide, but won't do it. Just need someone to talk to me.\\
Response: Suicide. Reasoning: The use of the phrase 'suicidal' immediately suggests that the person is struggling with suicidal ideation...\\ 

Post:...\\

Post: Guided Meditation. Disclaimer: I am in no way saying meditation should be the sole method of treating bipolar disorder. I wouldn't be stable...\\
Response: Bipolar disorder. Reasoning:
\end{tabular}\\\midrule
T-SID & \begin{tabular}[c]{@{}l@{}} You will be presented with a post that possibly shows mental disorders, and an assigned label to show the type of the mental disorder from the following causes list:\\ No mental disorders, Suicide or self-harm tendency, Depression, PTSD. You must explain the reasoning of the assigned label step by step. Here are some examples:\\

Post: I didn't kill myself bc of this song pic.twitter.com/kffmJTpubz\\
Response: Suicide or self-harm tendency. Reasoning: The use of the phrase 'I didn't kill myself' immediately suggests that the person is struggling with suicidal...\\

Post:...\\

Post: Sad to hear the news @dickc @Twitter. You have great and will be missed by all your friends and supporters @Unilever pic.twitter.com/HjSaN4mBmp\\
Response: No mental disorders. Reasoning:
\end{tabular}\\\midrule
SAD & \begin{tabular}[c]{@{}l@{}} You will be presented a post that shows stress, and an assigned label to show the cause of the stress from from the following stress causes\\ list: School, Financial problem, Family issues, Social relationships, Work, Health issues, Emotional turmoil, Everyday decision making,\\ Other causes. You must explain the reasoning of the assigned label step by step. Here are some examples:\\

Post: i have been wanting to find another job for some time now\\
Response: Work. Reasoning: The post explicitly mentions that the poster has been wanting to find another job for some time now. This indicates...\\

Post:...\\

Post: I am so tired I like can't wake myself up.\\
Response: Health issues. Reasoning: 
\end{tabular}\\\midrule
CAMS & \begin{tabular}[c]{@{}l@{}} You will be presented with a post that shows mental disorders, and an assigned label to show the cause of the mental disorders from the following causes\\ list: Bias or abuse, Jobs and career, Medication, Relationship, Alienation, None. You must explain the reasoning of the assigned label\\ step by step. Here are some examples:\\

Post: Any advice? I start studying soon and am wondering how on earth I'll be able to concentrate...\\
Response: None. Reasoning: The post does not suggest that the poster has experienced bias or abuse, job-related stress...\\

Post:...\\

Post: Punch bullies in their faces, prevent broken family relationships, be better equipped for the real world, stop abusive father, have friends.\\
Response: Bias or abuse. Reasoning: 
\end{tabular}\\\midrule
loneliness & \begin{tabular}[c]{@{}l@{}} You will be presented with a post and an assigned label to identify whether the poster suffers from loneliness. Consider the\\ emotions expressed from this post to explain the reasoning of the label step by step. Here are some examples:\\

Post: Today would have been my best friend's 18th birthday, we'd be going out together for the first time, we'd be sitting here making...\\
Response: Yes. Reasoning: The post mentions his best friend's 18th birthday, going out together for the first time, and making resolutions...\\

Post:...\\

Post: Looking back, I am literally in the same spot I was a year ago (physically, as well as mentally), at one of my best-friends houses...\\
Response: No. Reasoning: 
\end{tabular}\\\midrule
MultiWD & \begin{tabular}[c]{@{}l@{}} You will be presented with a post and an assigned label to identify whether the wellness dimension of spiritual exists in the post, according to\\ Dunn’s model of psychological wellness. You must consider these information to explain the reasoning of the label step by step. Here are some examples:\\

Post: I question my purpose daily. Will I ever find the 'one' for me? This spiral of emotions I deal with on a daily basis...\\
Response: Yes. Reasoning: In the post, the individual expresses their search for meaning and purpose in their existence by questioning...\\

Post:...\\

Post: Since the week started I missed out every single day of college. I just can't sleep on time... It's 4 AM right now and I've come to the sad conclusion...\\
Response: No. Reasoning:
\end{tabular}\\\midrule
IRF & \begin{tabular}[c]{@{}l@{}} You will be presented with a post an assigned label to identify whether the post shows risk of perceived burdensomeness, considering the interpersonal\\ risk factors of mental disturbance in the post. You must consider these information to explain the reasoning of the label step by step. Here are some examples:\\

Post: Do you ever sleep 17 hours? Help me not feel so alone\\
Response: No. Reasoning: The post itself does not directly indicate any feelings of burden or thoughts of being better off gone. It is simply asking...\\

Post:...\\

Post: My partner and I recently moved across country. I was doing so well up until now. No homesickness, not even really upset about leaving, found a job...\\
Response: No. Reasoning: 
\end{tabular} \\ \bottomrule
\end{tabular}}
\end{table*}

\end{document}